\documentclass{article}
\usepackage{spconf,epsfig}
\usepackage{amsmath,amsfonts}
\usepackage{multirow}
\usepackage{threeparttable}
\usepackage{booktabs}

\usepackage[square,sort&compress,numbers]{natbib}
\usepackage{textcomp}
\usepackage{stfloats}
\usepackage{url}
\usepackage{verbatim}
\usepackage{graphicx}
\usepackage{hyperref}
\usepackage{array}

\usepackage{amsmath}
\usepackage{amsfonts} 
\usepackage{amssymb} 
\usepackage{multirow}
\usepackage{threeparttable}
\usepackage{booktabs}
\usepackage{bbm, dsfont}
\usepackage{bbding}
\usepackage{ifsym}

\usepackage[table]{xcolor}

\title{Learning to detect cloud and snow in remote sensing images from noisy labels}
\name{{Zili~Liu$^{1,2}$, Hao~Chen$^2$, Wenyuan~Li$^3$, Keyan~Chen$^1$, Zipeng~Qi$^1$,
Chenyang~Liu$^1$, Zhengxia~Zou$^1$, Zhenwei~Shi$^{1,*}$}
\sthanks{The work was supported by National Natural Science Foundation of China (62125102), National Key Research and Development Program of China (2022ZD0160401), Beijing Natural Science Foundation (JL23005), Fundamental Research Funds for the Central Universities, National Key Research and Development Program of China (NO.2022ZD0160101). $^*$Corresponding author.}
\thanks{This work was partially done during his internship at Shanghai Artificial Intelligence Laboratory.}
\address{$^1$Beihang University, $^2$Shanghai Artificial Intelligence Laboratory, $^3$University of Hong Kong}}

\begin{document}
%
\maketitle
\begin{abstract}
Detecting clouds and snow in remote sensing images is an essential preprocessing task for remote sensing imagery. Previous works draw inspiration from semantic segmentation models in computer vision, with most research focusing on improving model architectures to enhance detection performance. However, unlike natural images, the complexity of scenes and the diversity of cloud types in remote sensing images result in many inaccurate labels in cloud and snow detection datasets, introducing unnecessary noises into the training and testing processes. By constructing a new dataset and proposing a novel training strategy with the curriculum learning paradigm, we guide the model in reducing overfitting to noisy labels.
Additionally, we design a more appropriate model performance evaluation method, that alleviates the performance assessment bias caused by noisy labels. By conducting experiments on models with UNet and Segformer, we have validated the effectiveness of our proposed method. This paper is the first to consider the impact of label noise on the detection of clouds and snow in remote sensing images.
\end{abstract}

\vspace{-10pt}
\section{Introduction}
\label{sec:intro}
\vspace{-7pt}
Remote sensing is an indispensable tool for earth observation, providing critical data for environmental monitoring, disaster response, and climate research. However, the presence of clouds and snow in satellite imagery often obscures surface features, posing significant challenges to data accuracy and utility. Clouds can mask underlying terrain, while snow cover, with its similar spectral properties, can be difficult to distinguish from clouds. Accurate detection and differentiation of these two elements are crucial for the integrity of remote sensing applications, from weather forecasting to resource management. Cloud detection is a crucial preprocessing task that assesses the usability of remote sensing images and evaluates the imaging quality of these remote sensing data \cite{liu2023cloud}. During the cloud detection process, ground snow cover, with its extremely high reflective brightness, is often easily mistaken for clouds, leading to false detections. Consequently, the simultaneous detection of clouds and snow in remote sensing images has been widely studied \cite{wu2021geographic}. 

To facilitate the effective automated detection of clouds and snow in remote sensing images, previous work has proposed a variety of methods, which can be primarily divided into traditional spectral threshold-based approaches and the more recently rapidly evolving deep learning-based methods \cite{liu2023cloud}. This paper primarily focuses on deep learning-based approaches. Deep learning-based methods for cloud and snow detection predominantly employ semantic segmentation models from the field of computer vision \cite{minaee2021image} to perform pixel-wise classification on the input remote sensing images. Early studies mostly focused solely on the detection of clouds in remote sensing images and directly utilized existing models such as FCN (Fully Convolutional Networks) \cite{wu2018utilizing,francis2019cloudfcn} or UNet \cite{wieland2019multi}. Subsequently, as highly reflective snow significantly impacts the effectiveness of cloud detection, many studies began to explore the task of simultaneously detecting clouds and snow in remote sensing imagery. Related research has also incorporated prior information such as altitude, latitude, and longitude as inputs to enhance the precision of cloud and snow detection  \cite{wu2021geographic}. In a manner akin to the evolution of semantic segmentation tasks, later improvements have predominantly focused on enhancements to model architectures. These advancements include improving feature fusion techniques to increase the precision of detecting cloud and snow edges \cite{yang2019cdnet}, as well as reducing the need for extensive annotation through the use of self-supervised or weakly-supervised learning methods \cite{zou2019generative,li2020accurate}.
While many improved methods have boosted the performance of cloud and snow detection, there has been scant work specifically addressing the unique characteristics and challenges of the cloud and snow detection task. One of the challenges, which has persisted yet remains largely unaddressed, is the significant issue of \emph{noisy labels} within the cloud and snow detection task. Clouds in remote sensing images exhibit a variety of shapes, and the boundaries of many thin clouds are often difficult to discern. This results in a significant introduction of noisy labels during dataset annotation, particularly in scenes with thin clouds or mixed cloud-snow coverages, which is a problem not present in natural image scenes where the edges of objects are typically very distinct. Such noisy labels not only introduce noise during the training process but also use them directly for testing can lead to unreliable results, especially in scenarios where cloud pixels are difficult to distinguish. Therefore, designing specialized methods to mitigate the problems caused by noisy labels, which are particularly challenging in cloud and snow detection, is critically important.

To address the aforementioned issues, drawing on methods from the study of learning with noisy labels \cite{song2022learning}, we focus our improvements on the reconstruction of cloud and snow detection datasets and the incorporation of a curriculum learning-based training approach \cite{wang2021survey} to alleviate the adverse effects brought about by noisy labels. Specifically, we split the existing dataset into clean and noisy subsets based on the difficulty of distinguishing cloud and snow areas in the remote sensing images. Subsequently, during the training process, we start with the clean subset and gradually incorporate samples from the noisy subset into the training set, guiding the model to prioritize learning from the clean samples and reduce the impact of the noisy ones. In addition, we have developed a specialized model performance testing method for scenarios with noisy labels. The specialized design of the training and testing procedures described above enables us to obtain a more stable and better-generalizing model for the task of cloud and snow detection, which is plagued by a large number of noisy labels. We employed two mainstream networks, the CNN-based UNet \cite{ronneberger2015u} and the Transformer-based Segformer \cite{xie2021segformer}, to validate the effectiveness of our proposed method for learning with noisy labels.

The primary contributions of this paper include: (1) Introducing the label noise issue to the task of cloud and snow detection in remote sensing images, a problem that has been widely present but not studied within this domain.
(2) Constructing a new dataset and evaluation method for cloud and snow detection with noisy labels, taking into account different cloud types and remote sensing scenarios. (3)Proposing a cloud and snow detection method with the curriculum learning paradigm, tailored to the characteristics of this task.
\begin{figure*}
\centering
\includegraphics[width=0.85\linewidth]{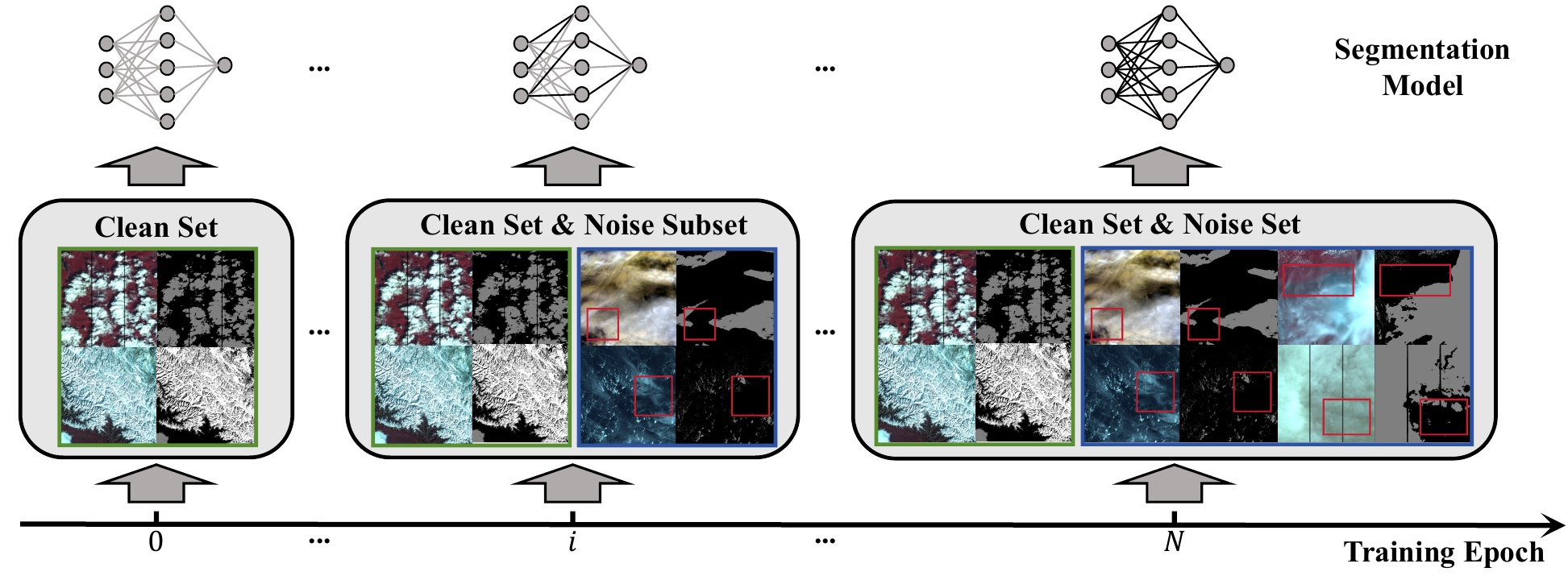}
\caption{Our proposed curriculum learning-based learning paradigm for cloud and snow detection from noisy labels. The red boxes represent the position of the noisy labels.}

\label{fig:overview}
\vspace{-14pt}
\end{figure*}
\vspace{-10pt}
\section{METHODOLOGY}
\vspace{-7pt}
\subsection{Overview}
\vspace{-5pt}

This section will analyze the issue of noisy labels in cloud and snow detection and propose a method to mitigate the negative impacts of noisy labels. As illustrated in Fig. \ref{fig:overview}, we divide the dataset into a clean set and a noisy set based on the difficulty of annotation. During the curriculum learning-based  \cite{wang2021survey} model training process, noisy samples are gradually introduced to alleviate the overfitting of noisy samples. The following sections will provide a detailed introduction to the construction of the dataset, improvements to the evaluation method, and the training process.
\vspace{-10pt}
\subsection{Dataset and Performance Evaluation with Noisy Labels}
\label{sec:benchmark}
\vspace{-5pt}
Based on our comprehensive study of publicly available cloud and snow detection datasets, we found that labels with noise are related to factors such as the type of clouds, the scenes in the remote sensing images, and whether there is a mixture of clouds and snow. Therefore, based on the aforementioned characteristics, we divided the original dataset into two parts: a \emph{clean set} and a \emph{noisy set}. The former primarily comprises samples with clear cloud boundaries, a homogeneous background, and a greater separation between clouds and snow, which typically have more accurate labels. The latter, on the other hand, mostly includes samples with fuzzy boundaries such as thin clouds, and a relatively complex background with more bright targets, or complicated situations with mixed clouds and snow.
Specifically, to ensure that our samples encompass a wide range of scenes and cloud types, we have selected approximately 2434 remote sensing images from various satellite sources. These include high-resolution satellites (GF-1), environmental monitoring satellites (HJ-2A/B), and commercial remote sensing satellites (SV2-01/02). Table \ref{tab:data} provides detailed information about the dataset.

\begin{table}[]
    \centering
    \renewcommand\arraystretch{1.5}
    \resizebox{0.9\linewidth}{!}{
    \begin{tabular}{c|cc|cc}
    \toprule
         \multirow{2}{*}{Subset Name} & \multicolumn{2}{c|}{Clean Set} & \multicolumn{2}{c}{Noisy Set}  \\
         & train \& valid & test & train \& valid & test\\
         \hline
         \# Images & 1185 & 450 & 599 & 200  \\
        \hline
         Data Source &  \multicolumn{4}{c}{Gaofen-1, SV2-01/02, HJ-2A/B} \\
        \hline
        Evaluation Metrics &  \multicolumn{2}{c|}{mIoU and OA} & \multicolumn{2}{c}{error\% $=\frac{\# images_{error}}{\# images_{total}}$}\\ 
    \bottomrule
    \end{tabular}}
    \caption{Description of the dataset setting for cloud and snow detection with noisy label.}
    \label{tab:data}
\end{table}

Since we have divided the samples into two subsets, specifically for the noisy set, the labels contain a considerable amount of noise, meaning the labels themselves are not accurate. Therefore, it is challenging to accurately evaluate the model's performance through specific numerical indicators during the testing phase. Furthermore, in practical applications, minor improvements in metrics are often insignificant. What is more critical is to avoid extensive false positives and false negatives on the scale of the entire image. In light of the issue with noisy labels, we have rethought the way we evaluate model performance. Specifically, as shown in Table \ref{tab:data}, we conduct different tests on the clean set and the noisy set. For the clean set, since the labels are relatively more accurate, we can directly utilize widely-used metrics such as Overall Accuracy (OA) and mean Intersection over Union (mIoU) \cite{liu2023cloud} as the primary evaluation criteria. As for the noisy set, we manually tallied the number of samples in the test results that exhibited extensive omissions and misclassifications. This includes the following three situations: (1) Misidentifying large areas of clouds as snow. (2) Overlooking large areas that should have been detected. (3) Wrongly detecting large areas where there shouldn't be any detection. By calculating the ratio of the number of samples that failed detection to the total number of samples, we can somewhat evaluate the model's generalization performance in complex scenarios.

\vspace{-10pt}
\subsection{Cloud-Snow Detection with Curriculum Learning}
\vspace{-5pt}

Based on the dataset we constructed, we have incorporated the curriculum learning paradigm into the training process of our cloud and snow detection model. To be specific, given the training pairs from the two datasets: $\mathcal{D}_{clean}$, $\mathcal{D}_{noisy}$ and the trainable parameters of the segmentation model $\Phi$. As shown in Fig. \ref{fig:overview}, at the beginning of the training phase, we use only the data from the clean set $\mathcal{D}_{clean}$ to train the model by:
\vspace{-5pt}
\begin{equation}
    \Phi_{epoch=0}^m := {\rm TrainStage1}(\forall\{x_i,y_i\}\in\mathcal{D}_{clean})
\end{equation}
which allows it to learn effectively under the supervision of accurate labels.
As the training progresses, with each epoch, we introduce data  with a proportion of $\frac{(epoch-m)}{(n-m)}*len(\mathcal{D}_{noisy})$ from the noisy set into the training set and continue the training process by:
\vspace{-5pt}
\begin{equation}
    \Phi_{epoch=m}^n := {\rm TrainStage2}(\forall\{x_i,y_i\}\in\mathcal{D}_{clean}\cup\mathcal{D}_{noisy}^{sub})
\end{equation}
where $m$ and $n$ represent the epoch numbers at which the integration of the noisy dataset begins and ends, respectively. This gradual infusion of noisy data is designed to help the model learn to cope with and generalize from the imperfections in the data. Finally, we train the model on the entire dataset, which includes both clean and noisy data by:
\begin{equation}
    \Phi_{epoch=n}^N := {\rm TrainStage3}(\forall\{x_i,y_i\}\in\mathcal{D}_{clean}\cup\mathcal{D}_{noisy})
\end{equation}
This three stages training approach allows the model to gradually adapt to noise while ensuring that it maintains its generalization performance and benefits from the increased sample volume.

\vspace{-10pt}
\section{Experiment}
\vspace{-7pt}
\subsection{Implementation Details}
\vspace{-5pt}
\subsubsection{Model Architecture}
\vspace{-5pt}
We employed two distinct models to ascertain the general applicability of our method. 
For the UNet model \cite{ronneberger2015u}, we utilized a ResNet-18 architecture as the encoder and connected features from its four stages to the decoder. The upsampling in the decoder is accomplished through a combination of bilinear interpolation and transposed convolution. Regarding the Segformer model \cite{xie2021segformer}, we employed a simplified version with a 2-layer decoder of 256 dimensions. Since the output dimensions of Segformer are a quarter of the input dimensions in both width and height, we performed a 4x upsampling on the output features using bilinear interpolation to ensure that the input and output dimensions are consistent.
\vspace{-7pt}
\subsubsection{Training Details}
\vspace{-5pt}
We used the Adam optimizer for both models and applied a step decay learning rate strategy. Considering the characteristics of CNN and Transformer models, we set the initial learning rates to 0.001 for the CNN-based UNet and 0.00006 for the Transformer-based Segformer. The learning rate is reduced by a factor of 10 every 10 epochs. For the loss function, we employed the standard cross-entropy loss and trained the models for a total of 150 epochs. Due to the limited number of samples in the dataset, we combined the training and validation sets and selected the model with the best mIoU on the training set for testing.

\vspace{-7pt}
\subsection{Results}
\vspace{-5pt}
Table \ref{tab:result} shows the results on the clean and noisy test set with our proposed curriculum learning-based training method compared to the original approach. We adhere to the evaluation approach described earlier in Section \ref{sec:benchmark}. Specifically, due to the substantial presence of noise in the labels of the noisy set, we directly computed the percentage ratio of the number of large-area detection errors to the total number of samples, instead of calculating specific numerical segmentation performance metrics. 
\begin{table}[]
    \centering
    \renewcommand\arraystretch{1.25}
    \resizebox{0.95\linewidth}{!}{
    \begin{tabular}{c|cccc|c }
    \toprule
         \multirow{2}{*}{Method} & \multicolumn{4}{c|}{Clean Set} & \multicolumn{1}{c}{Noisy Set}\\
         & $OA_{c}\uparrow$ & $OA_{s}\uparrow$ & $IoU_{c}\uparrow$ & $IoU_{s}\uparrow$ & error \% $\downarrow$\\
         \hline
         UNet \cite{ronneberger2015u}& 88.89 & 81.47 & 83.33 & 72.49 & 35 \\
         Segformer \cite{xie2021segformer}&  89.43 & 81.44 & 79.63 & 70.48 & 25.5 \\
         \hline
         UNet (Ours) & \textbf{92.56} & \textbf{85.86} & \textbf{87.50} & \textbf{78.77} & 23.5\\
         Segformer (Ours) & 89.44 & 78.81 & 84.84 & 73.94 & \textbf{18.5}\\
\bottomrule
    \end{tabular}}
    \caption{The comparison results between the original method and our proposed noise label learning method on two different model architectures.}
    \label{tab:result}
\end{table}

The comparative results indicate that after incorporating our proposed curriculum learning-based method, there are improvements in the metrics on both the clean and noisy test sets. It is important to note that a small improvement in numerical metrics on the clean test set does not necessarily translate to better generalization performance on the noisy test set. This implies that in practical applications, it is narrow-minded to solely pursue numerical improvements. Instead, it is crucial to ensure that the model can generalize well across different scenarios to minimize the occurrence of large-area detection errors. For instance, in the case of the Segformer, although its performance metrics on the clean set might be relatively inferior compared with UNet, it exhibits a stronger generalization capability in complex scenarios and across various cloud types on the noisy set, indicating a higher tolerance to noise. Moreover, integrating our proposed training strategy can further enhance the model's robustness to noisy labels.

\vspace{-15pt}
\section{Conclusion}
\label{sec:copyright}
\vspace{-7pt}
This paper is the first to highlight the problem of noisy labels for clouds and snow detection in remote sensing images. Starting from the curriculum learning paradigm, we constructed a dataset that separates clean data from noisy data. By progressively integrating noisy samples during the training process, we mitigated the issue of overfitting to noisy labels. In addition, we established a new model performance evaluation method suited for scenarios with noisy labels, aligning it more closely with the needs of practical applications. Experiments conducted on both a CNN-based UNet model and a Transformer-based Segformer model confirmed the effectiveness of our approach. This work lays the foundation for future research on learning with noisy labels in cloud and snow detection tasks.

\vspace{-15pt}
\small{
\bibliographystyle{IEEEbib}
\bibliography{refs}
}
\vspace{-15pt}
\end{document}